\documentclass{article}

\PassOptionsToPackage{numbers, compress}{natbib}

\usepackage[final]{nips_2018}

\usepackage[utf8]{inputenc} 
\usepackage[T1]{fontenc}    
\usepackage{hyperref}       
\usepackage{url}            
\usepackage{booktabs}       
\usepackage{amsfonts}       
\usepackage{nicefrac}       
\usepackage{microtype}      
\usepackage{color}
\usepackage{graphicx}
\usepackage{tikz}
\usepackage{mathpazo}
\usetikzlibrary{patterns}

\usepackage{booktabs}

\usepackage{amsmath}

\title{Intersectionality: Multiple Group Fairness in Expectation Constraints}

\author{
  Jack Fitzsimons \\
  Machine Learning Research Group \\
  University of Oxford, UK\\
  \url{jack@robots.ox.ac.uk}\\
  \And
  Michael Osborne \\
  Machine Learning Research Group \\
  University of Oxford, UK\\
  \url{mosb@robots.ox.ac.uk}\\
  \And
  Stephen Roberts \\
  Machine Learning Research Group \\
  University of Oxford, UK\\
  \url{sjrob@robots.ox.ac.uk}\\
}

\begin{document}

\maketitle

\begin{abstract}
Group fairness is an important concern for machine learning researchers, developers, and regulators. However, the strictness to which models must be constrained to be considered fair is still under debate. The focus of this work is on constraining the expected outcome of subpopulations in kernel regression and, in particular, decision tree regression, with application to random forests, boosted trees and other ensemble models. While individual constraints were previously addressed, this work addresses concerns about incorporating multiple constraints simultaneously. The proposed solution does not affect the order of computational or memory complexity of the decision trees and is easily integrated into models post training.
\end{abstract}

\section{Introduction}

The widespread use of machine learning algorithms and fully autonomous systems has greatly transformed the industrial landscape of the twenty-first century. However, with these great advances comes a responsibility for researchers, developers, and regulators to consider the impact of these systems on the broader society. In 2014, the US presidential administration published a report on big data collection and analysis, finding that ``big data technologies can cause societal harms beyond damages to privacy" \citep{united2014big}. The report raised the concern that algorithmic decisions inferred from big data may have harmful biases,  potentially leading to discrimination against disadvantaged groups.

This drive towards ethical practices in machine learning has led to many developments in \emph{algorithmic fairness}. One such advance has been towards developing algorithms which display group fairness, also referred to as statistical, conditional or demographical parity. From a regulatory viewpoint, group fairness is particularly interesting as affirmative action policies have already been passed to address discrimination against caste, race and gender \cite{weisskopf2004affirmative, dumont1980homo, deshpande2017affirmative}. However, it is worth noting there may be a considerable cost involved in achieving such fairness in some cases \cite{corbett2017algorithmic}.

In a machine learning context, there have primarily been two approaches towards developing systems which demonstrate group fairness; data alteration endeavors to modify the original dataset in order to prevent discrimination between groups \cite{luong2011k, kamiran2009classifying} in contrast to regularisation which penalizes models for unfair behavior \cite{kamishima2011fairness, berk2017convex, calders2013controlling, calders2010three, raff2017fair}.

More recently, there has been an effort towards constraining models such that they prohibit unfair behavior, a stricter assertion than regularisation. This work directly follows \cite{aistats_submission} in which group fairness in expectation for regression models is investigated, defined as:

\textbf{Group Fairness in Expectation (GFE):} \emph{A regressor $f(\cdot): X \rightarrow Y$ achieves group fairness in expectation with respect to groups $A, B \subseteq X$ and generative distributions $p_A(x)$ and $p_B(x)$ respectively iff,}
    \[ \mathbb{E}[f(x_i) | x_i \in A] - \mathbb{E}[f(x_i) | x_j \in B] = 0 \]
    \[ \int \left(p_A(x) - p_B(x) \right) f(x) dx =  0 \]
    
This work addresses an important issue raised in \cite{dwork2018group}, a model that satisfies conditional parity with respect to race and gender independently may fail to satisfy conditional parity with respect to the conjunction of race and gender. In the social science literature concerns about, potentially discriminated against, sub-demographics are referred to as intersectionality \cite{mccall2008complexity}. More formally, this work proposes a simple approach to ensure group fairness in expectation across an arbitrary set of subgroups. Applied to the popular decision trees, it is shown that provided the number of parity conditions is negligible compared to the number of training points, the order of computational and memory complexity is not increased. 

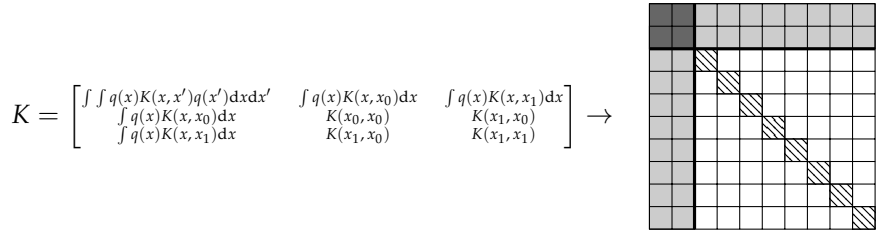
\begin{figure}[b]
\label{fig:kernel}
\begin{center}
\begin{minipage}[b]{0.45\linewidth}\centering
\[K = \begin{tiny}\begin{bmatrix}
    \int \int q(x) K(x, x') q(x') \textnormal{d}x \textnormal{d}x' & \int q(x) K(x, x_0)  \textnormal{d}x & \int q(x) K(x, x_1)  \textnormal{d}x\\
     \int q(x) K(x, x_0)  \textnormal{d}x & K(x_0, x_0) & K(x_1, x_0) \\
     \int q(x) K(x, x_1)  \textnormal{d}x & K(x_1, x_0) & K(x_1, x_1)\\
\end{bmatrix}\end{tiny} \rightarrow \]
\vspace{8mm}
\end{minipage}
\hspace{20mm}
\begin{tikzpicture}[scale=.30]
  \begin{scope}    
    \filldraw[fill=black!20!white] (1,8.) rectangle (10,10);
    \filldraw[fill=black!20!white] (0,0.) rectangle (2,9);
    
    \filldraw[fill=black!60!white] (0,8.) rectangle (2,10);


    \draw[pattern=north west lines, pattern color=black] (2,8.) rectangle (3,7);
    \draw[pattern=north west lines, pattern color=black] (3,7) rectangle (4,6);
    \draw[pattern=north west lines, pattern color=black] (4,6.) rectangle (5,5.);
    \draw[pattern=north west lines, pattern color=black] (5,5.) rectangle (6,4);
    \draw[pattern=north west lines, pattern color=black] (6,4) rectangle (7,3);
    \draw[pattern=north west lines, pattern color=black] (7,3.) rectangle (8,2.);
    \draw[pattern=north west lines, pattern color=black] (8,2.) rectangle (9,1);
    \draw[pattern=north west lines, pattern color=black] (9,1) rectangle (10,0);

    \draw (0, 0) grid (10, 10);
    \draw[very thick, scale=3] (3.333, 2.6666) -- (0, 2.6666);
    \draw[very thick, scale=3] (0.666, 0) -- (0.666, 3.333);
\end{scope}
\end{tikzpicture}

\end{center}
\caption{The above figure visualizes how the kernel matrix of a regressor can incorporate quadrature constraints. In the equations, $q(x)$ denotes the difference between two sub-populations $p_A(x)$ and $p_B(x)$. In the visual matrix representation (right) there are three components; the diagonal sub-matrix represents the leaves of the decision tree being independent of one another, the grey cells represent the relationship between the leaves and the constraints (denoted as $z$ in this work) and, finally, the dark upper left cells represents the interactions between constraints (denoted as $\bar{z}^T\bar{z}$ in this work).}
\end{figure}

\section{Constrained Kernel Methods}

As shown in \cite{aistats_submission}, kernel regressors may be constrained in terms of their expectation by adding auxiliary noiseless quadrature observations. Take for instance a Gaussian distribution with two dimensions. Given the distribution is zero mean, without loss in generality, correlation $\rho$ and variance  $\sigma_a^2$ and $\sigma_b^2$ respectively. With independent identically distributed noise $\sigma_n^2$, we can constrain the values of the expected outcomes by incorporating a noiseless observation on $\mu_a - \mu_b$ as follows,

\[
K = \begin{bmatrix}
    \sigma_a^2 + \sigma_b^2 - 2\rho\sigma_a\sigma_b & \sigma_a^2 - \rho\sigma_a\sigma_b & \sigma_b^2 - \rho\sigma_a\sigma_b\\
     \sigma_a^2 - \rho\sigma_a\sigma_b & \sigma_n^2 + \sigma_a^2 & \rho\sigma_a\sigma_b\\
     \sigma_b^2 - \rho\sigma_a\sigma_b& \rho\sigma_a\sigma_b& \sigma_n^2 +\sigma_b^2\\
\end{bmatrix}
\]

The above covariance matrix, $K$, has rank 2 with one perfectly observed value; namely the mean equality constraint. Thus inference on the two dimensions is constrained for $\sigma_n^2 >0$. 

As shown visually in Figure 1, this principle can be extended to Gaussian processes and other kernel regression techniques by using the differences in quadrature observations in order to incorporate mean equality constraints. Multiple constraints can also be created by simply adding more columns and rows to the kernel matrix accordingly.

\section{Constrained Decision Trees}

While the above, constrained kernel inference is interesting, the widespread impact comes into play when we extend the result to decision tree regression, random forests, boosted trees and other ensemble techniques. This is said not to dismiss the importance or value in kernel methods more broadly, but rather due to their popularity amongst \emph{data scientists}, a profession more common than machine learning researchers \cite{wu2008top}.  

While decision trees can be represented in either compressed or explicit kernel representation \cite{aistats_submission}, for the sake of conciseness this work will present results only for compressed representation. Thus, we will endeavor to minimize the perturbations induced on a per leaf bases, irrespective of the number of data points per leaf. The core difference between single and multiple constraints is that we can no longer use the arrowhead matrix lemma, instead, we must work out the update using the block matrix inversion lemma. Importantly, $p_A(x)$ and $p_B(x)$ for each constraint are defined as the empirical distributions of each subgroup considered. This is an important point as small subgroups may have empirical distributions which are not good approximations of the true generative distributions and hence our constrained space for inference may not constrain predictions to equate accordingly.

The kernel function in the compressed representation is simple the identity matrix, $K = \textnormal{I}$, that is to say each leaf is independent of one another. The kernel regression equation can be denoted as,

\[
f(\bar{\textnormal{x}}) = \mathbb{E}[\bar{\textnormal{y}}] = K_{\bar{x}, x} K_{x, x}^{-1} \textnormal{y}
\]

where the first $L$ values (number of leaves) of $K_{\bar{x}, x}$ indicate to which leaf $\bar{\textnormal{x}}$ belongs and the remaining values are set as zero, as the point to predict will not contribute to the empirical distributions of the subgroups under an inductive learning paradigm. 
 
\begin{figure}[b!]
\centering
  \includegraphics[width=\textwidth]{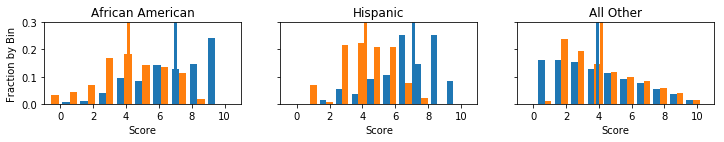}
  \caption{The figure shows the effect of GFE constraints in the inferred scores of the ProPublica dataset between African American, Hispanic and all other defendants. Before perturbation is in blue and after in orange. Vertical lines indicate the mean of the distributions.}
  \label{fig:demo1}
\end{figure} 
 
Using the block matrix inversion lemma we find,

\[K^{-1} = \begin{bmatrix}
    (1 + \sigma_n^2)\textnormal{I} & z\\
    z^T & \bar{z}^T\bar{z}\\
\end{bmatrix}^{-1} = {\tiny \begin{bmatrix}
    (1 + \sigma_n^2)^{-1} \textnormal{I} - (1 + \sigma_n^2)^{-2} z \left(\bar{z}^T\bar{z} - (1 + \sigma_n^2)^{-1} z^Tz \right)^{-1} z^T  & - (1 + \sigma_n^2)^{-1} z \left(\bar{z}^T\bar{z} - (1 + \sigma_n^2)^{-1} z^Tz \right)^{-1}\\
    - (1 + \sigma_n^2)^{-1} \left(\bar{z}^T\bar{z} - (1 + \sigma_n^2)^{-1} z^Tz \right)^{-1} z^T & \left(\bar{z}^T\bar{z} - (1 + \sigma_n^2)^{-1} z^Tz \right)^{-1} \\
\end{bmatrix}} \]

By simply inserting this into the kernel regression equation and noting that the elements of $\bar{z}$ are necessarily zero, the following update to the expected mean can be found as,

\[
f(\bar{\textnormal{x}}) = \frac{1}{1 + \sigma_n^2} \left( \textnormal{y}_j -  z_j \left(z^Tz \right)^{-1} z^T\textnormal{y} \right),
\]

with $z_j$ indicating the row of $z$ relating to the difference of subgroup distributions on leaf $j$. The effect of the noise can be removed by post multiplying by $1 + \sigma_n^2$.

\section{Experiments}

\subsection{ProPublica \& the COMPAS System}

The first experiment reproduces the experiment in \cite{aistats_submission} which uses a random forest to estimate the recidivism decile scores of the COMPAS algorithm applied to the ProPublica dataset while adding a GFE constraint between African Americans and Non-African Americans. However, it can also be noted that Hispanics also receive a similar discrimination. Figure 2, visualizes the effect of GFE constraints on the predicted distributions of the three demographics.  

\subsection{Illinois State Employee Salaries}

The Illinois state employee salaries\footnote{\url{https://data.illinois.gov/datastore/dump/1a0cd05c-7d17-4e3d-938d-c2bfa2a4a0b1}} since 2011 can be seen to have a gender bias and bias between veterans and non-veterans. The motivation of this experiment was if one wished to predict a fair salary for future employees based on current staff. Gender labels were inferred using the employees' first names, parsed through the \emph{gender-geusser} python library. GFE constraints were applied between all intersections of gender and veteran / non-veterans, the marginals of gender and the marginals of veteran / non-veterans. Table 1 shows the expected outcome of each group before and after GFE constraints are applied and Figure 3 visualizes the perturbations to the marginals of each demographic intersection due to the GFE constraints. The train-test split was set as 80\%-20\% and the incorporation of the GFE constraints increase the root mean squared error from \$12,086 to \$12,772, the cost of fairness. 

\begin{figure}[t!]
\centering
  \includegraphics[width=\textwidth]{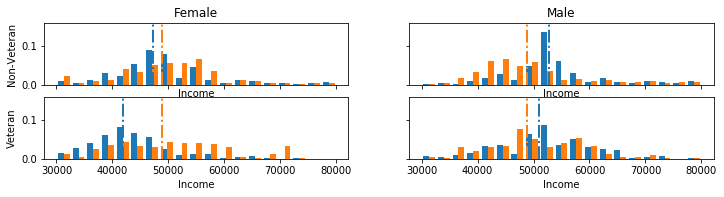}
  \caption{The figure visualizes the distribution of salaries before and after perturbations due to GFE constraints. It is clear that female veterans benefit the most from such a constraint, while male non-veterans lose out. Colors and lines denoting the same meaning as Figure 2. The figures are cropped to the main mode of salaries to facilitate visual comparisons.}
  \label{fig:demo2}
\end{figure}

\begin{table}[t!]
\hspace{-13mm}{\small
\begin{tabular}{c| c c c c || c c|| c c }
\toprule
Group& Female Non-Vet. & Male Non-Vet. & Female Vet. & Male Vet. & Male & Female & Vet. &Non-Vet.\\
\hline
Original&47,334&52,777&41,890&51,063&46,962&52,215&49,555&49,805\\
Perturbed&\emph{48,695} & \emph{48,693}& \emph{48,694} & \emph{48,693}&\emph{48,695}&\emph{48,698}&\emph{48,775}&\emph{48,775}\\
\bottomrule
\end{tabular}
}
\caption{The above table shows the expected outcome of a random tree regressor with and without GFE constraints applied to four sub-demographics, between gender and between veterans and non-veterans.}
\end{table}

\section{Conclusion}

Regulatory bodies have shown precedent in developing affirmative action and other group fairness policy. This work extends previous efforts to develop group fairness constrained machine learning techniques. While relatively simple to understand and easy to incorporate into models used by practitioners, the methodology of this paper has a direct impact to four of the ten top data science algorithms according to \cite{wu2008top}. All source code used in the experiments are available at \url{https://github.com/OxfordML/Fair_Regression.git}.

\medskip

\small

\bibliographystyle{abbrv}
\bibliography{bibfile}

\begin{thebibliography}{10}

\bibitem{berk2017convex}
R.~Berk, H.~Heidari, S.~Jabbari, M.~Joseph, M.~Kearns, J.~Morgenstern, S.~Neel,
  and A.~Roth.
\newblock A convex framework for fair regression.
\newblock {\em arXiv preprint arXiv:1706.02409}, 2017.

\bibitem{calders2013controlling}
T.~Calders, A.~Karim, F.~Kamiran, W.~Ali, and X.~Zhang.
\newblock Controlling attribute effect in linear regression.
\newblock In {\em Data Mining (ICDM), 2013 IEEE 13th International Conference
  on}, pages 71--80. IEEE, 2013.

\bibitem{calders2010three}
T.~Calders and S.~Verwer.
\newblock Three naive bayes approaches for discrimination-free classification.
\newblock {\em Data Mining and Knowledge Discovery}, 21(2):277--292, 2010.

\bibitem{corbett2017algorithmic}
S.~Corbett-Davies, E.~Pierson, A.~Feller, S.~Goel, and A.~Huq.
\newblock Algorithmic decision making and the cost of fairness.
\newblock In {\em Proceedings of the 23rd ACM SIGKDD International Conference
  on Knowledge Discovery and Data Mining}, pages 797--806. ACM, 2017.

\bibitem{deshpande2017affirmative}
A.~Deshpande.
\newblock Affirmative action in india.
\newblock In {\em Race and Inequality}, pages 77--90. Routledge, 2017.

\bibitem{dumont1980homo}
L.~Dumont.
\newblock {\em Homo hierarchicus: The caste system and its implications}.
\newblock University of Chicago Press, 1980.

\bibitem{dwork2018group}
C.~Dwork and C.~Ilvento.
\newblock Group fairness under composition.
\newblock {\em FATML}, 2018.

\bibitem{aistats_submission}
J.~Fitzsimons, A.~Al~Ali, M.~Osborne, and S.~Roberts.
\newblock Group fairness under composition.
\newblock {\em arXiv preprint arXiv:1810.05041}, 2018.

\bibitem{kamiran2009classifying}
F.~Kamiran and T.~Calders.
\newblock Classifying without discriminating.
\newblock In {\em Computer, Control and Communication, 2009. IC4 2009. 2nd
  International Conference on}, pages 1--6. IEEE, 2009.

\bibitem{kamishima2011fairness}
T.~Kamishima, S.~Akaho, and J.~Sakuma.
\newblock Fairness-aware learning through regularization approach.
\newblock In {\em Data Mining Workshops (ICDMW), 2011 IEEE 11th International
  Conference on}, pages 643--650. IEEE, 2011.

\bibitem{luong2011k}
B.~T. Luong, S.~Ruggieri, and F.~Turini.
\newblock k-nn as an implementation of situation testing for discrimination
  discovery and prevention.
\newblock In {\em Proceedings of the 17th ACM SIGKDD international conference
  on Knowledge discovery and data mining}, pages 502--510. ACM, 2011.

\bibitem{mccall2008complexity}
L.~McCall.
\newblock The complexity of intersectionality.
\newblock In {\em Intersectionality and Beyond}, pages 65--92.
  Routledge-Cavendish, 2008.

\bibitem{raff2017fair}
E.~Raff, J.~Sylvester, and S.~Mills.
\newblock Fair forests: Regularized tree induction to minimize model bias.
\newblock {\em arXiv preprint arXiv:1712.08197}, 2017.

\bibitem{united2014big}
{U}nited {S}tates. {E}xecutive Office of~the {P}resident and J.~{P}odesta.
\newblock {\em Big data: Seizing opportunities, preserving values}.
\newblock White House, Executive Office of the President, 2014.

\bibitem{weisskopf2004affirmative}
T.~E. Weisskopf.
\newblock Affirmative action in the {U}nited {S}tates and {I}ndia.
\newblock {\em A Comparative Perspective, London and New York: Routledge},
  2004.

\bibitem{wu2008top}
X.~Wu, V.~Kumar, J.~R. Quinlan, J.~Ghosh, Q.~Yang, H.~Motoda, G.~J. McLachlan,
  A.~Ng, B.~Liu, S.~Y. Philip, et~al.
\newblock Top 10 algorithms in data mining.
\newblock {\em Knowledge and information systems}, 14(1):1--37, 2008.

\end{thebibliography}

\end{document}